\pdfoutput=1
\documentclass[11pt,tables]{article}
\usepackage[table]{xcolor}
\usepackage{tablestyles}
\usepackage[]{ACL2023}
\usepackage{siunitx}
\usepackage{times}
\usepackage{latexsym}
\usepackage[T1]{fontenc}

\usepackage[utf8]{inputenc}

\usepackage{microtype}

\usepackage{inconsolata}

\title{Exploring Large Language Models for Classical Philology}

\author{Frederick Riemenschneider\\
  Dept. of Computational Linguistics\\
  Heidelberg University\\
 69120 Heidelberg\\
  \texttt{riemenschneider@cl.uni-heidelberg.de} \\\And
  Anette Frank \\
  Dept. of Computational Linguistics\\
  Heidelberg University\\
  69120 Heidelberg\\
  \texttt{frank@cl.uni-heidelberg.de} \\}

\usepackage{csquotes}
\usepackage{hyphenat}
\usepackage{arydshln}
\usepackage{amsmath}
\usepackage[greek,english]{babel}
\usepackage{tabularx}
\usepackage[]{cleveref}
\usepackage{calc}
\usepackage{tabularray}
\usepackage[inline]{enumitem}
\usepackage{graphicx}
\usepackage{multirow}
\usepackage{enumitem}
\usepackage{booktabs}
\usepackage{xspace}
\newcommand{\greBERTTa}{\textsc{Gr\textgreek{ε}(BERT$\mid$T)a}\xspace}
\newcommand{\tfiveboth}{\textsc{Phil(BERT$\mid$T)a}\xspace}

\newcommand{\greTa}{\textsc{Gr\textgreek{ε}Ta}\xspace}
\newcommand{\greBERTa}{\textsc{Gr\textgreek{ε}BERTa}\xspace}
\newcommand{\agbert}{AG \textsc{BERT}\xspace}
\newcommand{\tfive}{\textsc{T5}\xspace}
\newcommand{\bert}{\textsc{BERT}\xspace}
\newcommand{\roberta}{\textsc{RoBERTa}\xspace}
\newcommand{\mtfive}{\textsc{mT5}\xspace}
\newcommand{\electra}{\textsc{ELECTRA}\xspace}
\newcommand{\philta}{\textsc{PhilTa}\xspace}
\newcommand{\glem}{\textsc{GLEM}\xspace}
\newcommand{\philberta}{\textsc{PhilBERTa}\xspace}
\newcommand{\cltk}{\textsc{CLTK}\xspace}
\newcommand{\udpipe}{\textsc{UDPipe}\xspace}
\newcommand{\gretaenc}{\textsc{Gr\textgreek{ε}Ta-Enc}\xspace}

\usepackage{arydshln}

\Crefformat{footnote}{fn.\ #2#1#3}
\usepackage{tikz-dependency}
\usepackage{pifont}
\newcommand{\xmark}{\textcolor{red}{\ding{55}}}
\newcommand{\cmark}{\textcolor{green!20!black!70}{\ding{51}}}

\begin{document}

\maketitle
\begin{abstract}
Recent advances in NLP have led to the creation of powerful language models for many
languages including Ancient Greek and Latin. While prior work on Classical lan\-gua\-ges unanimously uses
BERT, in this work we create four language models for Ancient Greek that vary along two dimensions to study their versatility for tasks of interest for Classical languages: we explore (i) encoder-only and encoder-decoder architectures using \roberta and \tfive as strong model types, and create for each of them (ii) a monolingual Ancient Greek and a multi\-lin\-gu\-al instance that includes Latin and English.
We evaluate all models on 
morphological and syntactic tasks, including lemmatization, which demonstrates the added value of \tfive's
decoding abilities. 
We further define
two probing tasks to investigate the knowledge acquired by models pre-trained on Classical texts.
Our experiments provide the first benchmarking analysis of existing models of Ancient Greek.  Results show that our models provide significant improvements over the SoTA. The systematic analysis of model types can inform future research in designing language models for Classical languages, including the development of novel generative tasks.
We make all our models
available as community resources, along with a large
curated pre-training corpus for Ancient Greek, to support the creation of a larger, comparable model zoo for Classical Philology. Our models and resources are available
at \url{https://github.com/Heidelberg-NLP/ancient-language-models}.
\end{abstract}

\section{Introduction}
\label{sec:introduction}
Since the beginning of the creation of the \textit{Index Thomisticus} in 1946 \cite{thomisticum} and the publication of the Concordance to Livy \citep{concordance}, Classical Philology has been revitalized by the \enquote{digital revolution} \citep{berti2019digital}. Today, numerous efforts have been undertaken to make Classical texts digitally available, annotate, 
and automatically process them. E.g., the Classical Language Toolkit (\cltk, \citealp{johnson-etal-2021-classical}) offers various tools to process pre-modern languages, in particular Latin and pre-modern Greek.\footnote{
We use the term \enquote{pre-modern} and \enquote{Ancient} interchangeably
following the 
convention of calling every pre-modern language stage \enquote{Ancient}. This is in line with, e.g. \citet{singh-etal-2021-pilot}, \citet{plutarchsshadows}, and the ISO code standard.}

Recently, we see a surge of the first pre-trained contextualized language models (PLMs) for Classical languages:
Latin \textsc{BERT} has been proposed by \citet{bamman2020latin},
Ancient Greek (AG) \textsc{BERT} by \citet{singh-etal-2021-pilot}. Lately,
a second AG
\textsc{BERT} has been proposed by \citet{plutarchsshadows}. However, both AG \textsc{BERT} models have been pre-trained on a comparatively small pre-training dataset. Moreover, they have been initialized from Modern Greek \textsc{BERT} \cite{greekbert}, which limits them
to the modern Greek alphabet, ignoring the diacritics of Ancient Greek.

Although numerous richly annotated treebanks are available for Latin and AG, systems have, by now, not been evaluated on a shared benchmark. 
Given that
two popular treebanks for AG have been integrated into Universal Dependencies  \citep{de-marneffe-etal-2021-universal}, it is surprising that researchers
working on AG do not compare to benchmarking results of, e.g., 
\citet{straka-2018-udpipe}. Hence, a thorough assessment of the performance of the existing models is necessary in 
order to compare and evaluate their effectiveness for this underexplored language.

While \textsc{BERT} models are known to achieve high
performance on a wide range of tasks, encoder-decoder models or multilingual models may often be a better choice, depending on the task.
In this work, we explore a variety of language models for Classics in general and Ancient Greek in particular: We introduce \greTa, \greBERTa, \philberta, and \philta, four PLMs for Classics. \greBERTa and \greTa are \roberta \citep{roberta} and \tfive \citep{2020t5} models trained on Ancient Greek texts, respectively. \philberta and \philta are their trilingual counterparts pre-trained on Greek as well as Latin and English data.

We explore the advantages of \begin{enumerate*}[label=(\roman*)] \item the two model architectures in
\item mono- and multilingual
pre-training 
\end{enumerate*} for the mid-resource language Ancient Greek on a variety of morphological, syntactic, and semantic tasks, helping to answer questions, such as: \textit{When to choose one architecture over the other?} or: \textit{How does multilinguality affect a language model?} 

Moreover, we publish the first wide-ranging benchmark results to compare our models for AG and Latin to the relevant prior work, establishing new SoTA results for both languages.

In summary, we aim to unify and push forward the current research landscape at the intersection of Classics and NLP with the following contributions:

\begin{enumerate}[label=(\roman*), noitemsep]
 \item We introduce four pre-trained language models for Classics: \greBERTTa 
and \tfiveboth.  To 
our know\-ledge, we are  the first to develop encoder-de\-co\-der models for Classics, and multi\-lin\-gu\-al models tailored to both Latin and Greek.
    \item We evaluate the already existing
    and our proposed
    models on several tasks, making many of them comparable for the first time. Furthermore, we outperform the existing Ancient Greek \bert models by a notable margin.
    \item Our evaluation sheds light on the differences between encoders like \roberta and encoders of encoder-decoder models like \tfive as well as on the influence of multilinguality on the mid-resource language Ancient Greek.  By offering novel model types for AG, we aim to inspire new research and application tasks.  
    \item We develop and publish a large-scale, high-quality pre-training corpus for AG as a contribution to the community.
\end{enumerate}

\section{Related Work}
\paragraph{Pre-training Data for Ancient Greek.}
Pre-trained language models require large amounts of unlabeled 
pre-training data. 
Ancient Greek and Latin being
historical languages, the number of available texts is inherently limited, which 
makes the 
creation of a high-quality pre-training corpus even more important. To circumvent this problem, \citet{singh-etal-2021-pilot} and \citet{plutarchsshadows} 
pre-trained their AG \bert model  from a Modern Greek BERT \citep{greekbert}. 
But
this approach has two weaknesses: First, 
there is an important cultural 
gap between modern and ancient texts that we do not want to introduce into our models. A Modern Greek \bert is familiar with contemporary concepts like cell phones or communism, which are unknown to antiquity, while we intend to use PLMs as a \enquote{window} to ancient cultures.
Also 
the style of 
modern internet documents is fundamentally different from the
transmitted
ancient texts. Second, and more importantly, continuing pre-training of the Modern Greek \bert prevents us from adapting its
tokenizer. AG, however, 
uses more diacritics, which host
important information.  
By contrast, in our work, we build a tokenizer from scratch that is optimized for Ancient Greek. 

In order to boost the data needed to train \enquote{pure} models of Ancient Greek, we put special effort into the curation of a large, but high-quality 
pre-training corpus for AG, 
leveraging previously unused textual sources. Finally, 
we evaluate the effect of using additional multilingual pre-training data.

\begin{figure*}[!th]
\resizebox{\linewidth}{!}{
\includegraphics[width=\linewidth]{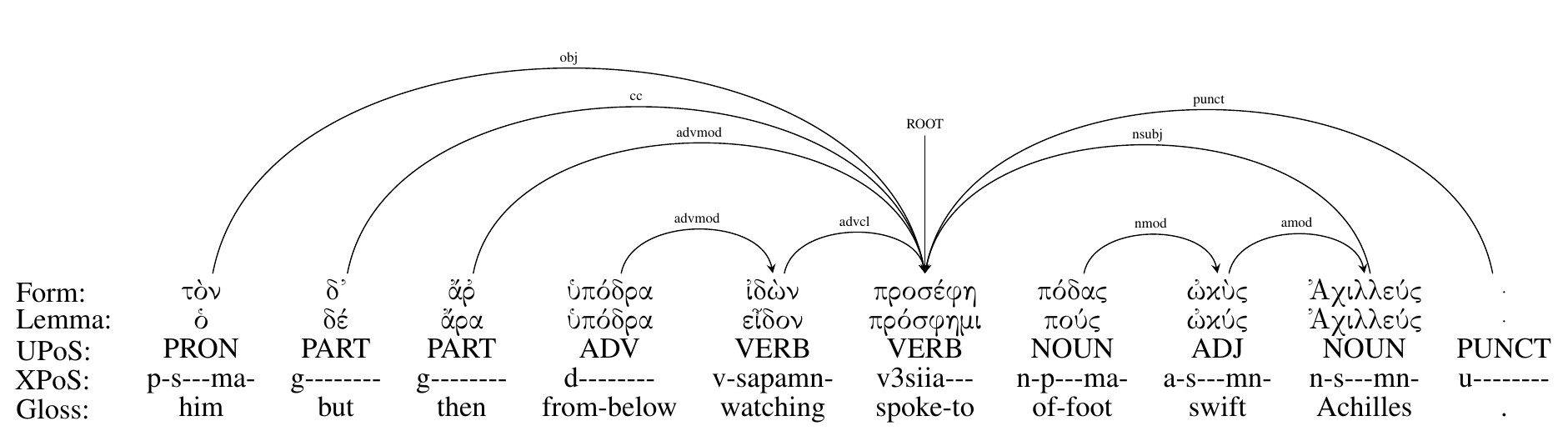}}
\caption{Example sentence (Hom.~\textit{Il}. 1.148) with corresponding dependency, lemma, UPoS, and XPoS labels. Translation: \enquote{Then watching him grimly from under his brows, swift-footed Achilles spoke to him.}}
\label{fig:example}
\end{figure*}

\paragraph{Evaluating Models for Ancient Languages.}

Morphological and syntactic tasks, such as PoS tagging, dependency parsing, and lemmatization, have always
been of interest to researchers of Latin and Ancient Greek. The standard tool for AG morphological analysis is Morpheus \citep{morpheus}, a rule-based system, that has also been integrated into many more recent approaches. 
PoS Tagging has also been performed by various language-agnostic systems
trained on AG data \citep{CelanoCraneMajidi+2016}, but their success depends heavily on the chosen dataset: a winning system on one dataset 
\citep{CelanoCraneMajidi+2016} achieves the worst results on another \citep{Keersmaekers}.
More recently, the \cltk \citep{johnson-etal-2021-classical} provides a variety of taggers for many tasks.
Surprisingly, 
although
numerous richly annotated treebanks are available, systems have, by now, not been
evaluated on a common
benchmark.\footnote{Cf. also \citet{johnson-etal-2021-classical}: \enquote{The
\cltk lacks formal evaluations of its models’ accuracies. [...] Unfortunately, [outside] benchmarks do not yet exist for pre-modern languages.}}
E.g., \citet{singh-etal-2021-pilot}
test their proposed AG \bert on random splits from three popular treebanks, which we cannot compare against. The second  AG \bert \citep{plutarchsshadows} has only been evaluated on authorship attribution.  

As for lemmatization,  \citet{LemmatizationforAncientGreek} provide an evaluation of three different lemmatizers. However, one of the evaluated candidates was partly trained on test data, which may have influenced its performance.
It is noteworthy that, despite the integration of two popular treebanks for AG into Universal Dependencies (UD, \citealp{de-marneffe-etal-2021-universal}), many groups working on AG systems have not compared their models against the results of models benchmarked on UD data,
such as \citet{straka-2018-udpipe}. We remedy these issues
by evaluating our 
systems and existing AG \bert models on the two 
authoritative treebanks covered by UD. The tasks we consider -- dependency parsing, lemmatization, coarse, universal (UPoS) PoS tagging and fine-grained, language-specific (XPoS) tagging -- are visualized in \Cref{fig:example}.

For Latin, the issue does not arise thanks to the EvaLatin 2022 campaign \citep{sprugnoli-etal-2022-overview}, which has enabled direct comparison of models and has engendered strong models for Latin.
Yet, despite the impressive results achieved in EvaLatin, our trilingual models outperform the existing systems on PoS tagging and lemmatization.

\paragraph{Language Model Architectures.}
Language models can 
be categorized into three classes: encoder-only, decoder-only, and encoder-decoder models. Encoder-only models such as \bert \citep{devlin-etal-2019-bert} and \roberta \citep{roberta} are best suited for tasks that aim to analyze
complete sequences by sequence or token classification. Encoder-decoder models, on the other hand, are typically employed for conditional generation tasks, such as machine translation.
Currently, all three models for ancient languages are \bert and thus encoder-only architectures.

We argue that an encoder-decoder model, such as \tfive \citep{2020t5}, is a useful addition to this encoder-only landscape. First, it enlarges the space of possible NLP tasks for AG, enabling us, e.g., to cast
lemmatization as a sequence-to-sequence task and to explore machine translation for ancient languages. Second, it allows us to compare the encoder of an encoder-only model with that of an encoder-decoder architecture, as they are both trained on the same data with a similar pre-training objective. Finally, commonly used multilingual encoder-decoder models like \textsc{mT5} \citep{xue-etal-2021-mt5} and \textsc{ByT5} \citep{xue-etal-2022-byt5} are not pre-trained on Ancient Greek texts.

As we aim for optimally trained encoder-only models, we chose
\roberta over \bert: 
its dynamic masking strategy exploits the 
pre-training data better, and it has been shown that \bert's NSP objective can be 
detrimental \citep{roberta}. 

\section{Pre-trained Language Models for Ancient Greek and Latin}
\subsection{\greBERTTa and \tfiveboth}

\greBERTa and \philberta are \textsc{RoBERTa}$_{\text{base}}$-, \greTa and \philta are \textsc{T5}$_{\text{base}}$-sized models. Both models are pre-trained using a masked language modeling (MLM) objective. Specifically, in the case of \roberta, wordpieces are masked during the pre-training process, while for \tfive, spans are masked. 
Although it has been shown that multilingual pre-training can lead to gains for low-resource languages through cross-lingual transfer, it remains an open question when exactly it is preferable to use
a multilingual 
instead of a monolingual model \citep{primer}. To explore the implications of multilinguality for AG
language models, we test different capabilities and possible interferences by comparing the different model types.

\subsection{PLM Fine-tuning for Downstream Tasks\footnote{The following 
descriptions remain neutral to different PLM types by referring to basic transformer components. Where necessary, we will distinguish specific PLM types.}}
\label{subsec:methodology}
\paragraph{PoS Tagging.}
PoS tagging for 
Ancient Greek
typically aims for
a complete morphological analysis: Next to the word class,
the model has to predict eight fine-grained morphological attributes.\footnote{Person, number, tense, mood, voice, gender, case, and degree. Usually, many of these attributes are left empty. E.g., only adjectives have the attribute \enquote{degree}.} We frame this sequence labeling task as a multi-task classification problem applied to each token, with nine different classification heads per token on top of one shared encoder: We denote a sequence of tokens $S$ of length $n$ as $S = w_1, w_2,\dots,w_n$ and 
refer to the
contextualized embedding of each token
as $\mathbf{e}_i = \text{Encoder}(w_{1:n}, i)$.
As Byte Pair Encoding \citep{sennrich-etal-2016-neural} splits words into sub\-word units, we represent each token using its first subword embedding in the encoder. Each of the nine att\-ri\-butes is predicted using a feed-forward network app\-lied to the last encoding layer, followed by a soft\-max function.  The total loss is 
calculated as:
\[
\mathcal{L}_{\text{total}} = \sum_{m=0}^{8} \frac{1}{9} \mathcal{L}_m
\]
We use this approach for the 
Perseus XPoS dataset. For the other, less-detailed tagsets, we employ a single classification head.

\paragraph{Dependency Parsing.}
We follow \citet{zhang-etal-2017-dependency} who cast
dependency parsing as head selection. The model predicts a unique head for each token considered as a dependent. Since the model makes independent predictions, the obtained dependency graph 
can (in a few cases) be unconnected and is then completed by the Chu-Liu-Edmonds algorithm \citep{chuliu,edmonds} for building non-projective trees -- given that AG allows free word order. While \citeposs{zhang-etal-2017-dependency} \textsc{DeNSe} parser was based on a bi-directional
LSTM, we define the model on top of the final hidden states of the transformer encoders.

Following \citet{zhang-etal-2017-dependency}, we add an artificial \textsc{Root} token $w_0$ and calculate the probability of the word $w_j \in \{ w_0, w_1, \dots , w_N \}$ being the head of the word $w_i \in \{w_1, w_2,\dots, w_n \}$ in $S$ as:
\[
p_{\text{head}} (w_j | w_i, S) = \frac{\exp({f(\mathbf{e}_j, \mathbf{e}_i)})}{\sum_{k=0}^{N} \exp({f(\mathbf{e}_k, \mathbf{e}_i)})}
\]
where $f$ predicts the score of an edge $(w_j, w_i)$ as follows: 
\[
f(\mathbf{e}_j, \mathbf{e}_i) = \mathbf{v}^\top \cdot \tanh(\mathbf{U} \cdot \mathbf{e}_j + \mathbf{W} \cdot \mathbf{e}_i)
\]
Here, $\mathbf{v}$ is a weight vector and $\mathbf{U}$, 
$\mathbf{W}$ 
weight matrices. Dependency labels are predicted in a similar fashion: Let $g$ 
be a single hidden-layer rectifier network that takes as input the concatenation \([\mathbf{e}_i; \mathbf{e}_j]\). The probability for the label
$l$ is then computed as:
\[
p_{\text{label}} (l | w_j, w_i, S) = \frac{\exp({g(\mathbf{e}_j, l, \mathbf{e}_i)})}{\sum_{l' \in L} \exp({g(\mathbf{e}_j, l',\mathbf{e}_i)})}
\]
While \citet{zhang-etal-2017-dependency} use the representations of their trained \textsc{DeNSe} model as input for the label classifier, we resort to the pre-trained embeddings.

\paragraph{Lemmatization.}
Current systems for lemmatization of AG, such as
\udpipe \citep{straka-2018-udpipe} or \glem \citep{glem}, are rule-based or use a classifier to predict editing rules that
modify a token's pre- and suffixes. However, these complex scripts are not well-suited for a language like AG, which has many irregular forms that involve modifications of the word stem.
An alternative approach is to utilize an encoder-decoder model 
that receives
the inflected form, the PoS tag, and (optionally) additional information such as morphological features, as demonstrated for different languages by \citet{schmid} or \citet{wrobel-nowak-2022-transformer}.

Yet, these earlier encoder-decoder-based lemmatization models are purely word-based and rely on pre-computed PoS tags or morphological features in a pipeline setting. By contrast, we propose a novel \textsc{T5}-based lemmatization model that is (i) \textit{contextualized}, so that relevant morphological indicators can be inferred by the model on the fly from the token's surrounding context. (ii) The model works \textit{end-to-end}: it receives the surface form of the word to be lemmatized in its full sentence context and predicts its lemma without receiving or predicting PoS tags or morphological features.\footnote{However, multi-task learning for joint morphological analysis \textit{and} lemmatization is an interesting option that we did not pursue here.} 
We mark the t(arget) token to be lemmatized in its context using 
delimiter tokens 
\texttt{$<$t\_tok\_beg$>$} and \texttt{$<$t\_tok\_end$>$}.
For instance, for the input sentence 
\textgreek{ξύνοιδα} \texttt{$<$t\_tok\_beg$>$} \textgreek{ἐμαυτῷ} \texttt{$<$t\_tok\_end$>$} \textgreek{οὐδὲν ἐπισταμένῳ} with the marked inflected t(arget) token
\textgreek{ἐμαυτῷ}, we expect as output the lemma \textgreek{ἐμαυτοῦ}. 
We also experiment with providing,  in addition, the target word as a sequence of individual characters, delimited by an additional separator token \texttt{$<$t\_tok\_sep$>$}:
\textgreek{ξύνοιδα} \texttt{$<$t\_tok\_beg$>$} \textgreek{ἐμαυτῷ} \texttt{$<$t\_tok\_sep$>$} \textgreek{ἐ μ α υ τ ῷ} \texttt{$<$t\_tok\_end$>$} \textgreek{οὐδὲν ἐπισταμένῳ}. 

\paragraph{Semantic and World Knowledge Probing Tasks.} \label{par:semantics}
So far, we considered only morphological and syntactic tasks. However, to evaluate the models more comprehensively, it is necessary to also test their semantic and world knowledge. Since such benchmarks do not exist for AG or Latin, we create two small datasets to evaluate these aspects. Inspired by \citet{talmor-etal-2020-olmpics}, we test whether the language models can \textbf{distinguish synonyms from antonyms}. For this task, we input a sentence, e.g., \textgreek{τὸ χρήσιμον καὶ τὸ ἀγαθόν:} $<$mask$>$ \textgreek{ὁμοῖά ἐστιν} (\enquote{the useful and the good: they are $<$mask$>$ similar}), and the model has to predict either \textgreek{οὐχ} (\enquote{not}) or \textgreek{πάντως} (\enquote{very}). 
\citet{talmor-etal-2020-olmpics} cast a similar task for English as a zero-shot MLM prediction problem using 
\bert and \roberta. However, 
with our prompt, the models always predict \textgreek{οὐχ} (\enquote{not}), regardless of 
the provided word pairs.
Experiments with variations of the prompt have led to similar difficulties. Hence, we evaluate this task in a few-shot setting, fine-tuning the MLM-head on
10 to 50 shots of synonyms and antonyms each, to 
prepare them for the task.

Similarly, we construct a dataset of
\textbf{family relationships} between (mythical) heroes and gods. Here,
the model is given a phrase, such as \textgreek{Τηλέμαχος ὁ τοῦ} $<$mask$>$ \textgreek{παῖς} (\enquote{Telemachus, son of $<$mask$>$}), and has to predict the correct entity (in this case, Odysseus). For this task, we test the models in 
a zero-shot setting.
However, this task cannot be solved by most
encoder-only models, as the masked names typically consist of more than a single wordpiece.
Thus, for this task, we evaluate only \greTa and \philta, 
which can predict full entity names. By comparing the mono- and multilingual variants, we assess the models' 
acquired world knowledge as well as potential effects that may be induced by multilingual training: Given that Greek and Roman mythology share many of these gods, yet by different names, the multilingual model may be able to acquire additional knowledge from the Latin pre-training data, to solve the task formulated in Ancient Greek.
We describe both datasets in \Cref{app:datasetcreation}.

\subsection{Acquisition of Pre-training Data}

\paragraph{Ancient Greek.} 
To cover a wide range of dialects, topics, and time periods of Ancient Greek, we make use of 
four different data sources: (the Greek part of) the Open Greek \& Latin Project,\footnote{\url{https://opengreekandlatin.org/}.} the \textsc{CLARIN} corpus Greek Medieval Texts,\footnote{\url{https://inventory.clarin.gr/corpus/890}.} the Patrologia Graeca,\footnote{\url{https://patristica.net/graeca/}.} and the Internet Archive (IA).\footnote{\url{https://archive.org/}.}
While the first three sources contain born-digital 
textual data, the IA online library 
provides books in the public domain along with their OCR
transcriptions.

However, we found the partition of texts labeled as Ancient Greek in the IA 
to be incomplete and noisy:
only a small fraction of the books containing AG text was labeled as such, 
and only few of them were transcribed with OCR settings 
supporting Greek characters. 
We hence extracted a novel data partition from the IA that was then fully re-OCRed by the Internet Archive to ensure correct transcription. To select
a large number of high-qua\-li\-ty texts, we applied
a complex retrieve and filter procedure, focusing
not only on (i) text quality, but also on (ii) collecting purely Ancient Greek texts, avoiding inclusion of texts in different languages, such as Latin, English, or German that can co-occur in the same book, and on (iii) filtering duplicates. 

\paragraph{Latin and English.} Acquiring pre-training data for Latin  was facilitated by the Corpus Corporum pro\-ject,\footnote{\url{https://www.mlat.uzh.ch/}.} a meta-repository 
of 
Latin corpora 
that offers a comprehensive representation of the Latin language. All this data was kindly offered to us.

For English, we collect pre-training data from various sources, aiming for texts that 
are
related to antiquity, by being focused on
the same topics that we find in
ancient texts -- as opposed to 
modern themes. 
To this end, we utilize English translations of Latin and Ancient Greek texts as pre-training data. Furthermore, we ensure that the amount of English data is of similar size as the ancient texts,  to prevent the models from being overwhelmed by a large number of English texts.

Statistics of pre-training data 
in \Cref{tab:statistics}. More details on corpus creation and     statistics in \Cref{app:corpuscreation}.

\begin{table}
\centering
\resizebox{\linewidth}{!}{
\begin{tabular}{llr}
\theadstart

\textbf{Language} &
\textbf{Dataset} &
   \textbf{Number of Tokens}\\
\tbody

  \multirow{5}{*}{Ancient Greek} & Open Greek \& Latin & \hphantom{0}$30.0$ million\\
  &Greek Medieval Texts & \hphantom{00}$3.3$ million\\
   &Patrologia Graeca & \hphantom{0}$28.5$ million\\
   &Internet Archive & $123.3$ million\\
   &Overall& \textbf{185.1 million}\\
   \hdashline
   Latin & Corpus Corporum & \textbf{167.5 million}\\
   \hdashline
   English & Overall & \textbf{212.8 million}\\
   
 \tend
\end{tabular}}
\caption{Statistics of the pre-training datasets. Only Open Greek \& Latin is used by \citet{singh-etal-2021-pilot} and \citet{plutarchsshadows} for their
AG \bert models. Token counts 
determined by UNIX command \texttt{wc -w}.}

\label{tab:statistics}
\end{table}

\subsection{Pre-training Process}

Even though our filtering of the IA corpus resulted in high-quality texts, the corpus is necessarily noisier than the born-digital texts. We therefore
start pre-training on the IA data, and continue with the born-digital texts. Our tokenizers and the multilingual variants are trained on the born-digital texts only. For further pre-training details, see \Cref{app:pretraining}.

\section{Experiments}

We run the experiments outlined in \Cref{subsec:methodology} to provide insight into the performances achieved by different model types and in relation to prior SoTA.

\begin{table*}[ht!]
\centering
\resizebox{\linewidth}{!}{
\begin{tabular}{p{.1875\linewidth}p{.24\linewidth}p{.24\linewidth}p{.2\linewidth}p{.2\linewidth}p{.2\linewidth}}
\theadstart
& \multicolumn{2}{c}{\cellcolor{gray!25}\textbf{PoS Tagging}} & \multicolumn{2}{c}{\cellcolor{gray!25}\textbf{Dependency Parsing}} & \cellcolor{gray!25}\textbf{Lemmatization} \\
\tsubheadstart \cellcolor{gray!25} \tsubhead& \tsubhead\cellcolor{gray!25}UPoS & \tsubhead\cellcolor{gray!25}XPoS & \tsubhead\cellcolor{gray!25}Unlabeled & \tsubhead\cellcolor{gray!25}Labeled & \\
\tbody
Task Description & PoS tagging with universally applicable, coarse PoS tags
& PoS tagging with lang\-uage-\-specific, fine-grained 
tags; complete morphological analysis in the case of Perseus & predicting the head of each token in text & predicting the head and relation type of each token in text & predicting the lemma of each token in text\\
\cellcolor{gray!15}Metric & \cellcolor{gray!15}Accuracy  & \cellcolor{gray!15}Accuracy  & \cellcolor{gray!15}UAS & \cellcolor{gray!15}LAS & \cellcolor{gray!15}Accuracy \\
Datasets & \parbox{\widthof{EvaLatin}}{Perseus} \cmark \newline \parbox{\widthof{EvaLatin}}{PROIEL} \cmark \newline EvaLatin \cmark &\parbox{\widthof{EvaLatin}}{Perseus} \cmark \newline \parbox{\widthof{EvaLatin}}{PROIEL} \cmark \newline EvaLatin \xmark&\parbox{\widthof{EvaLatin}}{Perseus} \cmark \newline \parbox{\widthof{EvaLatin}}{PROIEL} \cmark \newline EvaLatin \xmark&\parbox{\widthof{EvaLatin}}{Perseus} \cmark \newline \parbox{\widthof{EvaLatin}}{PROIEL} \cmark \newline EvaLatin \xmark&\parbox{\widthof{EvaLatin}}{Perseus} \cmark \newline \parbox{\widthof{EvaLatin}}{PROIEL} \cmark \newline EvaLatin \cmark\\
\cellcolor{gray!15} Model Architecture & \cellcolor{gray!15}Encoder + Classification Head & \cellcolor{gray!15}Encoder + Classification Head(s) & \cellcolor{gray!15} Encoder + \textsc{DenSe} & \cellcolor{gray!15} Encoder + \textsc{DenSe} & \cellcolor{gray!15}Encoder-decoder\\
PLM Instances & (\textsc{Gr\textgreek{ε}}$\mid$\textsc{Phil})\textsc{BERTa}, (\textsc{Gr\textgreek{ε}}$\mid$\textsc{Phil})\textsc{Ta-Enc} & (\textsc{Gr\textgreek{ε}}$\mid$\textsc{Phil})\textsc{BERTa}, (\textsc{Gr\textgreek{ε}}$\mid$\textsc{Phil})\textsc{Ta-Enc} & (\textsc{Gr\textgreek{ε}}$\mid$\textsc{Phil})\textsc{BERTa}, (\textsc{Gr\textgreek{ε}}$\mid$\textsc{Phil})\textsc{Ta-Enc} & (\textsc{Gr\textgreek{ε}}$\mid$\textsc{Phil})\textsc{BERTa}, (\textsc{Gr\textgreek{ε}}$\mid$\textsc{Phil})\textsc{Ta-Enc} & (\textsc{Gr\textgreek{ε}}$\mid$\textsc{Phil})\textsc{Ta}\\\tend
\end{tabular}}
\caption{Summary of the tasks under consideration.}
\label{tab:overview}
\end{table*}

\subsection{Datasets}

\paragraph{Ancient Greek.}
For the PoS tagging, dependency parsing, and lemmatization tasks, we evaluate the PLMs for AG on the data provided by the
Perseus and the PROIEL datasets, which
are both integrated into Universal Dependencies 2.10 \citep{de-marneffe-etal-2021-universal}. 

To probe our models for semantic and world knowledge (see \Cref{par:semantics}), we use our newly constructed datasets, described in \Cref{app:datasetcreation}.

\paragraph{Latin.} For Latin, we resort to the treebank used in
EvaLatin 2022 \citep{sprugnoli-etal-2022-overview}, which covers
three tasks: PoS tagging, lemmatization, and feature identification. Since no data for dependency parsing is provided, we restrict our evaluation to 
PoS tagging and lemmatization. 
In EvaLatin, instead of constructing test data by drawing samples from the initial data set,
the test data exhibits different degrees of distribution differences in relation to the training data.
For each task, three 
test sets are provided: 
The \textit{Classical} set 
belongs to the same genre and time period as the training data, but comes from an author not included in the training data.
The \textit{Cross-genre} data includes two works that belong to different genres, yet being written during roughly the same time period. 
The \textit{Cross-time} test set is based on
text written in the 15th century, which is significantly later than the texts of
the training data. 

In \Cref{tab:overview}, we summarize the diverse tasks
under consideration with their corresponding metrics, the used evaluation datasets, the model architectures, and the pre-trained language models that are applicable to the respective task.
Further details, including dataset statistics, are provided in \Cref{app:udandevalatin}.

\subsection{Models and Baselines}
\paragraph{Ancient Greek.} To showcase the capabilities of a recent system tailored to AG, we report the results of the taggers provided by the Classical Language Toolkit \citep{johnson-etal-2021-classical}.\footnote{\label{fn:cltk}
From the multiple taggers offered by the \cltk we choose the one
that achieved best performance on the validation set. For the Perseus dataset, 
this is a TnT tagger \citep{brants-2000-tnt}, while for PROIEL, it is    Stanza \citep{qi2020stanza}.  Note, however, that it is not possible to directly compare the results to those of the other models, as they were trained on different data splits and using an older version of the dataset (cf.\
\url{https://github.com/cltk/greek_treebank_perseus).}}  As a
baseline, we use the currently best-performing system, \udpipe \citep{udpipe}, a transformer-based multitask architecture that utilizes multilingual \bert, trainable word embeddings, and character embeddings.\footnote{We report scores of the most recent, unpublished version of \udpipe 
(\url{https://ufal.mff.cuni.cz/udpipe/2/models\#universal_dependencies_210_models}) and the scores obtained when training \udpipe ourselves.}  In addition, to directly assess the benefits of using our monolingual model, we replace this multilingual \bert with our \greBERTa model. 

For PoS tagging and dependency parsing, we further compare to
both prior encoder models trained on AG texts.
We use the 
PoS tagger and 
\textsc{DeNse} 
(\Cref{subsec:methodology}) to 
evaluate both AG
\bert models as well as our \greBERTa and \philberta models.
We apply the same approach to \textsc{GreTa}'s encoder  (\textsc{GreTa-Enc}) to investigate its behavior.

For lemmatization, we compare the performance of \cltk and \udpipe with that of our full-fledged \tfive models. To predict a lemma during inference, we use beam search with a 
width of 20.

\paragraph{Latin.} For Latin, we report the results of both teams that participated in the EvaLatin 2022  competition:  Team \textsc{Kraków} \citep{wrobel-nowak-2022-transformer}
utilizes the \textsc{XLM-RoBERTa}$_\text{large}$ \citep{conneau-etal-2020-unsupervised} model for PoS tagging, team \textsc{KU-Leuven} \citep{mercelis-keersmaekers-2022-electra} employs an \electra model. For lemmatization, \citet{wrobel-nowak-2022-transformer} use \textsc{ByT5}$_\text{small}$ \citep{xue-etal-2022-byt5}, a multilingual encoder-decoder model
similar to \mtfive \citep{xue-etal-2021-mt5} that operates on UTF-8
bytes instead of subwords. \citet{mercelis-keersmaekers-2022-electra} implement a cascaded approach that resembles the Greek lemmatizer \textsc{GLEM} \citep{glem}: If a rule-based lookup returns multiple lemmata, the system tries to disambiguate between these possibilities by means of the predicted PoS tag.  To further clarify any remaining ambiguities, a classifier is trained to select the correct lemma from the available options.

\begin{table*}[]
\centering
\resizebox{0.8\linewidth}{!}{
\begin{tabular}{lllll}
\theadstart
   \cellcolor{gray!25}\textbf{Model} &
    \multicolumn{2}{c}{\cellcolor{gray!25}\textbf{PoS Tagging}} &
    \multicolumn{2}{c}{\cellcolor{gray!25}\textbf{Dependency Parsing}}\\
 \tsubheadstart
\cellcolor{gray!25} &
    \cellcolor{gray!25}\tsubhead UPoS &
    \cellcolor{gray!25}\tsubhead XPoS &
    \cellcolor{gray!25}\tsubhead UAS & 
    \cellcolor{gray!25}\tsubhead LAS 
    \\
\tbody
 \cltk & $68.83$ & $47.21$ & $59.21$ & $43.24$\\
\cdashline{1-5}
\udpipe (official) & $92.88$ & $85.60$& $80.32$ & $74.53$\\
\udpipe (ours) & $92.36$ $(0.09)$ & $84.72$ $(0.06)$& $78.74$ $(0.04)$ & $73.14$ $(0.06)$\\ 
 \udpipe + \greBERTa & $95.74$ $(0.06)$ & $90.95$ $(0.07)$& $86.30$ $(0.14)$& $82.15$ $(0.14)$\\
    AG \bert \cite{singh-etal-2021-pilot}& $94.92$ $(0.18)$& $88.27$ $(0.27)$& $84.03$ $(0.12)$& $78.80$ $(0.37)$\\
    AG \bert \cite{plutarchsshadows}& $92.50$ $(0.03)$& $84.56$ $(0.13)$& $80.34$ $(0.11)$& $74.22$ $(0.21)$\\
    \gretaenc & $94.44$ $(0.14)$& $89.03$ $(0.13)$ & $87.32$ $(0.04)$& $83.06$ $(0.07)$\\
    \philberta & $95.60$ $(0.21)$& $90.41$ $(0.18)$& $86.99$ $(0.06)$& $82.69$ $(0.06)$\\
    \greBERTa & $\mathbf{95.83}$ $\mathbf{(0.10)}$ & $\mathbf{91.09}$ $\mathbf{(0.02)}$ & $\mathbf{88.20}$ $\mathbf{(0.11)}$ & $\mathbf{83.98}$ $\mathbf{(0.21)}$\\
 \tend
\end{tabular}
}
\caption{PoS tagging and dependency parsing results on the Ancient Greek Perseus dataset. The results are
averaged over three runs with different random seeds, and the standard deviation is indicated in parentheses, except
for the \cltk and \udpipe (reported results). Note also that the \cltk is not trained on exactly the same data as
the other models and therefore not strictly comparable.}
\label{tab:posanddeps}

\end{table*}

\section{Results}

\paragraph{Ancient Greek.}
We present the results for 
\textbf{PoS tagging} and \textbf{dependency parsing} for Ancient Greek 
on the Perseus dataset in \Cref{tab:posanddeps}. 
The PROIEL dataset seems
to be easier to solve, as all models achieve performances that are much closer to each other
(see \Cref{app:furtherresults}). 
Since the overall trends are 
consistent across both datasets, we focus our discussion on the results on the Perseus dataset.

As seen in \Cref{tab:posanddeps}, the
\cltk performs clearly below
all 
other models on both tasks. While the
\cltk is not directly comparable
to the other models (see \Cref{fn:cltk}),
the evaluation still provides
a perspective on the capabilities of the \textit{de facto} only available framework for processing AG text.

\udpipe provides a strong baseline, which AG \bert \citep{plutarchsshadows} is unable to consistently outperform. By contrast, 
all other PLMs show clear gains over \udpipe. The monolingual, encoder-only \greBERTa  model consistently performs best on all tasks.
Interestingly, the performance of \textsc{GreTa-Enc} on PoS tagging is slightly worse than that of \philberta, while it achieves better results for dependency parsing. This trend has also been observed in initial experiments. We elaborate on the behavior of \gretaenc and \philberta in \Cref{sec:discussion}.

Results for \textbf{Lemmatization}
are shown
in \Cref{tab:lemmatization}. Here, augmenting \udpipe with \greBERTa's pre-trained embeddings does not lead to better scores. We attribute this to the tokenization process and refer to our discussion in \Cref{sec:discussion}. \greTa, on the other hand, demonstrates strong
encoder-decoder capabilities and significantly outperforms \udpipe. Providing \greTa with the individidual characters of the target word leads to a small gain.

\begin{table}
\centering
\resizebox{0.70\linewidth}{!}{
\begin{tabular}{ll}
\theadstart
      \textbf{Model} &
     \textbf{Accuracy}\\
\tbody
\cltk & $76.10$\\
\cdashline{1-2}
\udpipe (official) & $86.70$\\
\udpipe (ours) & $84.50$ $(0.09)$\\
\udpipe + \greBERTa & $85.56$ $(0.06)$\\
\philta & $90.02$ $(0.02)$\\
\philta + Chars & $90.66$ $(0.01)$\\
\greTa & $90.80$ $(0.10)$  \\
\greTa + Chars & $\mathbf{91.14}$ $\mathbf{(0.10)}$\\
 \tend
\end{tabular}
}
\caption{Lemmatization results for Ancient Greek on the Perseus dataset. Results are averaged over three runs, with 
standard deviation 
in parentheses, except for the \cltk and \udpipe (reported results).
}
\label{tab:lemmatization}
\end{table}

\begin{figure}
    \centering
    \includegraphics[width=\linewidth]{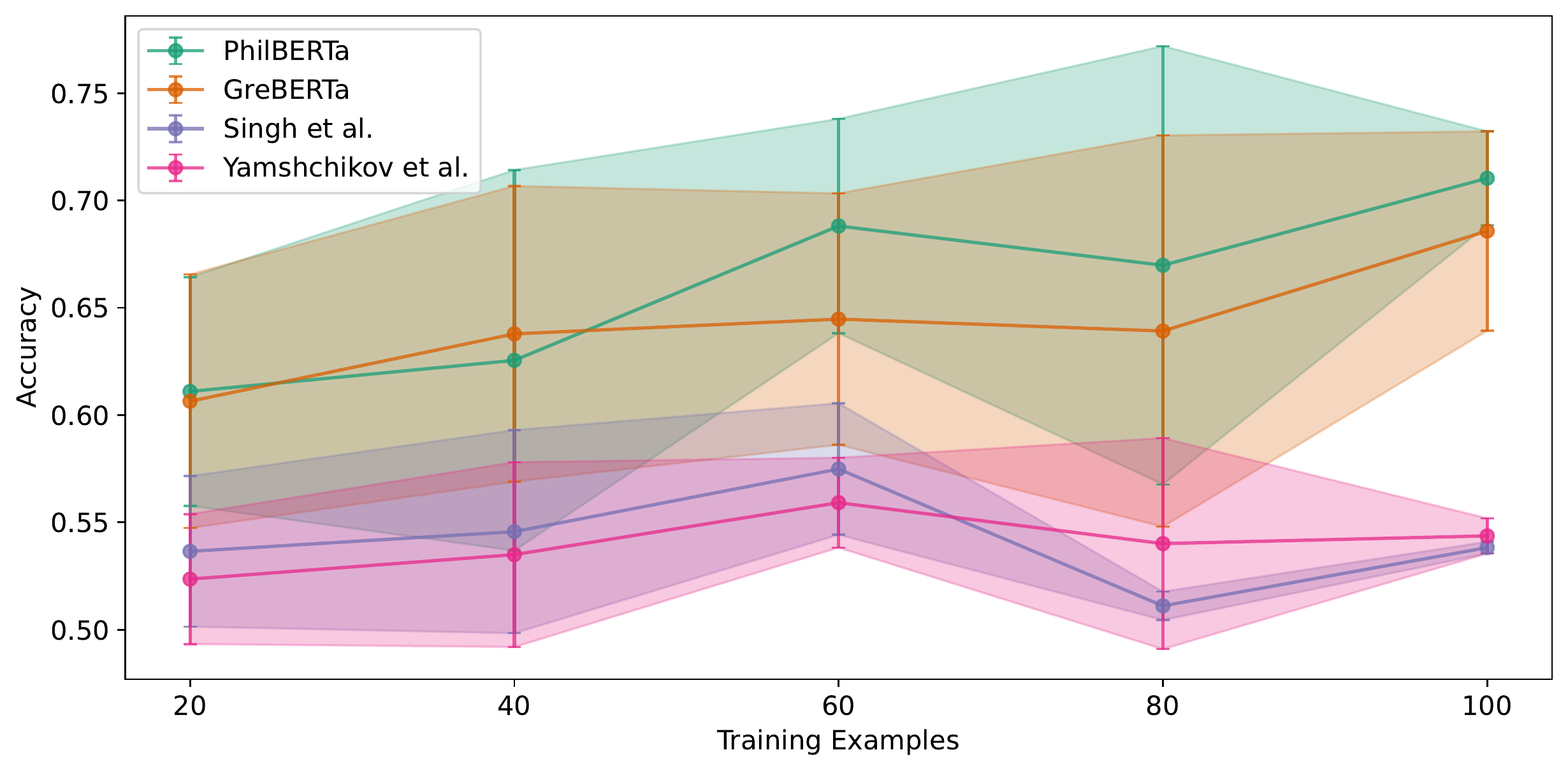}
    \caption{Synonym/antonym disambiguation accuracy  
    for growing
    few-shot 
    sizes: \greBERTa and \textsc{Phil}\-BERT\textsc{a} 
    vs.\ AG BERT models. 
    We use
equal amounts of synonyms and antonyms 
(a run with $20$ samples includes 
$10$ synonyms and $10$ antonyms). 
We use $k$-fold cross-validation. Error bars show standard deviation.}
    \label{fig:synonyms}
\end{figure}

The results of the 
\textbf{Synonym/antonym disambiguation task} are visualized in \Cref{fig:synonyms}. Again, \greBERTa and \philberta demonstrate higher scores compared to the \agbert models. We observe the same for \greTa and \philta (cf.\ \Cref{fig:synonymsencdec} in \Cref{app:furtherresults}). 
Our monolingual models and their multilingual counterparts perform almost equally, especially taking into account the overlapping standard deviation bands. We see a minimal trend for \philta to gain over \greTa in \Cref{fig:synonymsencdec}, but our small datasets do not allow drawing firm conclusions on their relative performance.

\begin{table}
\centering
\resizebox{0.8\linewidth}{!}{
\begin{tabular}{lllll}
\theadstart
\textbf{Model}&$\mathbf{k=1}$ & $\mathbf{k=5}$& $\mathbf{k=10}$
&$\mathbf{k>10}$\\
\tbody
\greTa & 4.39 & 9.65 & 10.53 & 10.96\\
\philta & 3.07 & 8.33 & 11.40 & 11.84 \\
 \tend
\end{tabular}}
\caption{Zero-shot family relationships task (recall@k).}
\label{tab:recallknowledge}
\end{table}

Finally, we report zero-shot results for the \textbf{Family relationship task} in \Cref{tab:recallknowledge}. As the \tfive-based models have been pre-trained to 
predict multiple masked spans at once, they tend to  
predict, for each sample, more than a single entity. We interpret
the output as a ranked list and report recall@k, 
evaluating whether the correct entity 
is contained in the first 1, 5, 10, and $>$10 predictions, restricting the maximum sequence length to 50 wordpieces.

\paragraph{Latin.} The PoS tagging and lemmatization scores on EvaLatin 2022 are reported in \Cref{tab:evalatinpos}. While the performance scores of all models are rather
close to each other, 
our trilingual models consistently outperform the EvaLatin 2022 participant systems 
across all three subtasks. 
\philberta reaches even higher scores than \textsc{Krak{\'o}w-open} on PoS tagging, which leverages additional annotated data.
On lemmatization,
\philta similarly outperforms \textsc{Krak{\'o}w-closed} on the Classical, Cross-genre, and Cross-time subtasks by 2.25, 1.78, and 0.23 percentage points, respectively, but does not outperform
\textsc{Krak{\'o}w-open} on the Cross-genre and the Cross-time subtask.

\begin{table}[t]
\centering
\resizebox{\linewidth}{!}{\begin{tabular}{lllll}
\theadstart
    &
      \textbf{Model} &
     \textbf{Classical} &
     \textbf{Cross-genre} &
     \textbf{Cross-time} \\
\tbody
\multirow{4}{*}{\rotatebox{90}{UPoS}} & \textsc{Krak{\'o}w-open} & $97.99$ \hphantom{$(0.06)$}& $96.06$ \hphantom{$(0.15)$}& $92.97$ \hphantom{$(0.12)$}\\
\cdashline{2-5}
&\textsc{Krak{\'o}w-closed} & $97.61$ \hphantom{$(0.06)$}& $94.62$ \hphantom{$(0.15)$}& $92.70$ \hphantom{$(0.12)$}\\
&\textsc{KU-Leuven} & $96.33$ \hphantom{$(0.06)$}& $92.31$ \hphantom{$(0.15)$}& $92.11$ \hphantom{$(0.12)$}\\
&\philberta & $\mathbf{98.23}$ $\mathbf{(0.06)}$ & $\mathbf{96.59}$ $\mathbf{(0.15)}$&$\mathbf{93.25}$ $\mathbf{(0.12)}$\\
\hline
\multirow{4}{*}{\rotatebox{90}{Lemmatiz.}}&\textsc{Krak{\'o}w-open} & $97.26$ \hphantom{$(0.06)$}& $96.45$ \hphantom{$(0.15)$}& $92.15$ \hphantom{$(0.12)$}\\
\cdashline{2-5}
&\textsc{Krak{\'o}w-closed} & $95.08$ \hphantom{$(0.06)$}& $91.62$ \hphantom{$(0.15)$}& $91.68$ \hphantom{$(0.12)$}\\
&\textsc{KU-Leuven} & $85.44$ \hphantom{$(0.06)$}& $86.48$ \hphantom{$(0.15)$}& $84.60$ \hphantom{$(0.12)$}\\
&\philta + Chars& $\mathbf{97.33}$ $\mathbf{(0.04)}$ & $\mathbf{93.40}$ $\mathbf{(0.13)}$&$\mathbf{91.91}$ $\mathbf{(0.04)}$\\
 \tend
\end{tabular}}
\caption{PoS tagging and lemmatization results (EvaLatin 2022 dataset). \textsc{Krak{\'o}w-open} uses
additional data.} 
\label{tab:evalatinpos}
\end{table}

\section{Analyses and Discussion}
\label{sec:discussion}

\paragraph{Training Behavior of \gretaenc.}
While \textsc{GreTa-Enc} and \greBERTa are of  similar size (\Cref{tab:pretrainingdetails}) and pre-trained with comparable objectives, \gretaenc performs slightly worse than \greBERTa. One reason may be that in a \tfive model, some important information is distributed across encoder and decoder.
This raises the question of whether encoders in encoder-decoder models are  
trained suboptimally, and whether improvements may be obtained by combining separately pre-trained encoders and decoders, or by pre-training the encoder before adding the decoder.
Another reason may be
that the encoder is not accustomed 
to using  its classification head. Here again, it may be advantageous to pre-train the encoder before extending it to encoder-decoder pre-training. 

\begin{figure}
    \centering
    \includegraphics[width=0.85\linewidth]{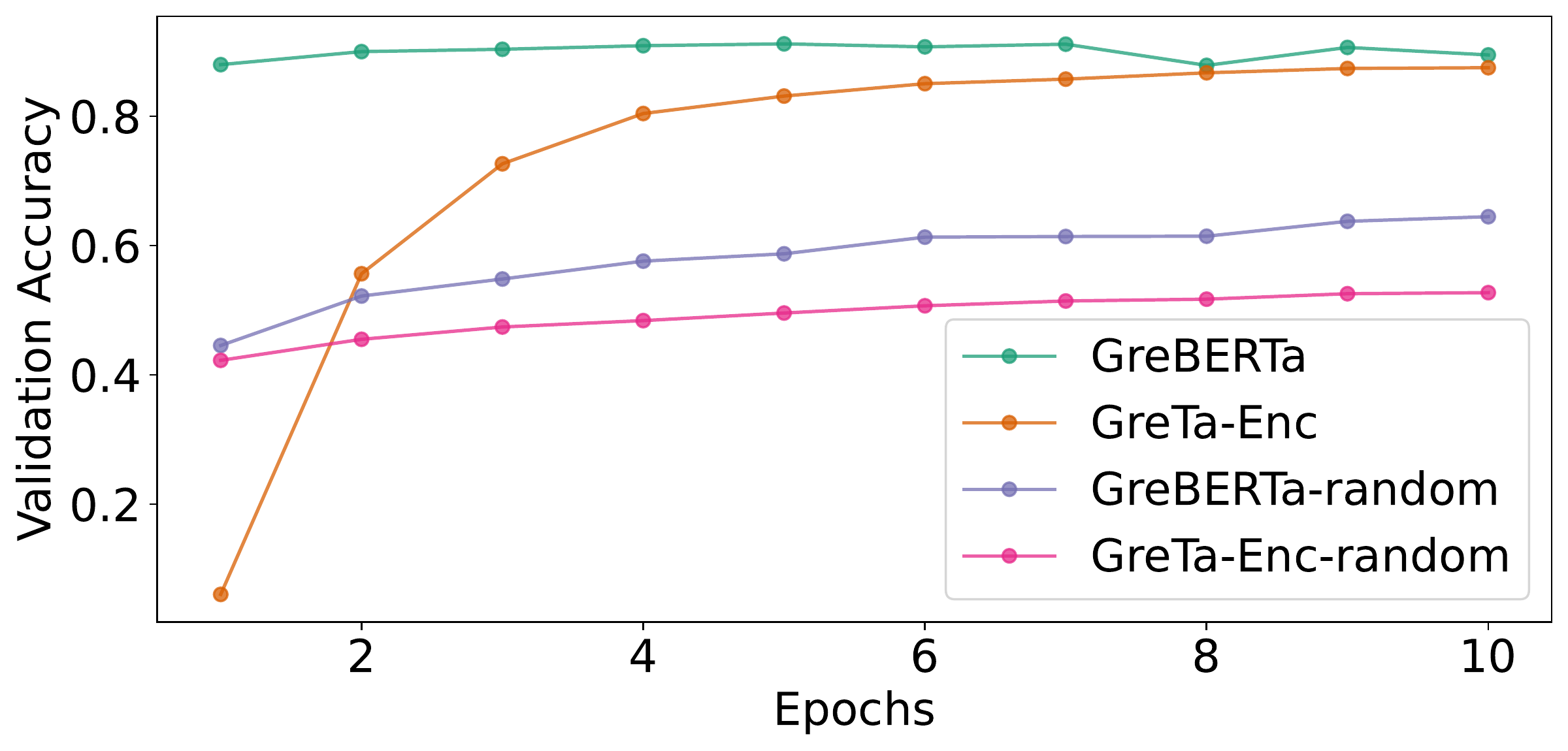}
    \caption{Validation accuracy (AG XPoS Tagging on Perseus) for 
    \gretaenc and \greBERTa 
    and 
    ran\-dom\-ly initialized counterparts 
    over various training epochs.}
    \label{fig:t5enc}
\end{figure}

In \Cref{fig:t5enc} we compare the PoS tagging validation accuracy of \gretaenc to that of a randomly initialized \tfive encoder (same size). 
\gretaenc performs much worse than the randomly initialized model after one epoch, reaching only approximately $6\%$. However, while the randomly initialized model stagnates, \gretaenc outperforms the randomly initialized model
after two epochs, significantly improving its performance thereafter. By contrast, \greBERTa reaches a high validation accuracy already after one epoch. We see the same trend
with different random seeds and for dependency parsing,
but it
is most apparent in Perseus XPoS tagging.

\paragraph{Lemmatization as a Character-based Task.}
As seen in \Cref{tab:lemmatization}, augmenting \udpipe with \textsc{Gr\textgreek{ε}}\-\textsc{BERTa}
does not lead to significant improvement for lemmatization. This we attribute to the tokenization process. \greBERTa uses wordpieces, which contain little information about individual characters. We hypothesize that
\udpipe 
ignores the \greBERTa embeddings for lemmatization and instead relies on its own additional character embeddings. Accordingly,
explicitly providing \greTa with the individual characters of the inflected word form 
leads to a slight increase in performance.

This explanation can also shed light on the success of the \textit{UTF-8 bytes-based} \textsc{ByT5} model 
for lemmatization in Latin. 
This model was chosen by \citet{wrobel-nowak-2022-transformer}, after initial experiments with the \textit{wordpiece-based} \textsc{mT5} \citep{xue-etal-2021-mt5} 
had underperformed. 
Future work on (AG) lemmatization could therefore investigate whether Byte Pair Encoding-based
models can be augmented with character embeddings as additional input.

\paragraph{Effect of Multilinguality.}
\Cref{tab:posanddeps} shows
that 
\philberta consistently performs slightly worse compared to monolingual \greBERTa on morphological and syntactic tasks. We attribute this to the \textit{curse of multilinguality} \citep{conneau-etal-2020-unsupervised}: 
the 
capacity of the trilingual models is split between three languages. Still,
both models
achieve strong results on 
AG and Latin tasks and can be especially useful in tasks that require multilingual knowledge, 
as in MT or glossing tasks. 
Our small-sized knowledge probing tasks show
very similar performance for both model types. While
the size of our data does not allow for firm conclusions, this is
in line with \citet{kassner-etal-2021-multilingual}, who find no   improved knowledge representation in multilingual PLMs. 

\section{Conclusion}
We introduce four strong language models for Classical Philology, including
the first encoder-decoder PLMs for Ancient Greek and Latin. 
We ri\-go\-rous\-ly benchmark our models and prior work on various 
tasks,  demonstrating strong
improvement over the SoTA. 
We demonstrate
the versatility of encoder-decoder models,  
(i) by offering a novel 
end-to-end contextualized lemmatization model for AG and Latin, with a  greatly simplified 
architecture 
that clearly
outperforms prior work; 
(ii) 
while MLM in encoder-only models is  
restricted to single-token predictions, 
our T5-based models exhibit great flexibility for formulating 
probing tasks, 
which help exploring what models learn from pre-training data.
Considering the two investigated model dimensions, 
our work (iii) sheds light on differences between the encoders of \tfive vs.\ \roberta, where the former tends to exhibit slower learning curves; (iv) 
our monolingual models outperform the multilingual ones
in 
monolingual morphological and syntactic tasks, without 
clear trends on small-scale semantic and knowledge probing tasks.

\section*{Limitations}
While we aim  for a comprehensive analysis of existing methods (such as lemmatization) and model types for Ancient Greek and other Classical languages, there are limits to exhaustively exploring the full space of variations and rigorously evaluating their impact on model performance. 
For example, we could not comprehensively evaluate the effects of (i) the pre-training corpora, as we did not re-train a BERT model for Ancient Greek, to pin down the exact difference between prior BERT models (which were trained on smaller data before) and our own models, which are based on inherently stronger model types; similarly, we did not induce Latin \roberta and \tfive models, to confirm the differences between mono- and multilingual models for language-specific Latin tasks. (ii) In a similar vein, we did not compare different model sizes. However, we studied prior work and scaling laws and believe that the base model is appropriate for the size of our training data. Further factors of this type concern (iii) hyperparameter settings and (iv) other factors in isolation. 

Not only do we miss sufficient computational resources to perform such manifold ablations and comparative assessments, we also considered the carbon footprint that such experiments cause and which does not stand up to the insights that could possibly be gained from more experiments. 

For these reasons, we focused on two selected dimensions of variants that we believe to be valuable for a community interested in Classical languages: 

(i) We tried to answer questions as to when multilingual models can be profitably used, and (ii) aimed to showcase various potential advantages of encoder-decoder models, which by now have not been considered in studies on Classical languages.  

Another clear limitation lies in the size of the demonstrated semantic and knowledge probing tasks.
(i) They are of small size, and we cannot, therefore, draw firm conclusions as to, e.g., the effect of multilinguality. Also, the synonym/antonym disambiguation task is presumably the most difficult one. As a counter-balance, we used a more tangible task for knowledge probing, by choosing family relationships, which we expect to be frequently found in the pre-training corpora.

(ii) A further limitation we find for the knowledge probing tasks resides in the size of our trained models and the underlying pretraining data. This limitation could be one that is not easy to overcome. But we still encourage the community to create similar probing task datasets. Future work may find appropriate ways of data augmentation, or transfer learning methods that are applicable to historical languages so that further progress and insight will be  possible.

\section*{Ethics Statement}
It is a computationally demanding task to pre-train large language models. However, transfer learning opens the possibility to fine-tune our pre-trained models, which showed strong performances, in a reasonable amount of time.  

The texts utilized for pre-training the models may well exhibit biases related to ancient perspectives of the world. We do not view this as an issue, as the proposed language models for historical languages are intended for academic use and do not have practical, everyday applications.

\section*{Acknowledgments}
We are deeply indebted
to the Internet Archive team for their continuous support by creating new OCR transcriptions of the misclassified Greek books, and to our anonymous reviewers for their comments, which have helped to significantly improve the paper. We thank Nina Stahl and Thomas Kuhn-Treichel for their help in creating our semantic and knowledge probing tasks, as well as Jan Ctibor and Philipp Roelli for providing us with the invaluable \textit{Corpus Corporum} data. Finally, we  acknowledge and thank for crucial support from 
the Google TPU Research Cloud program, for granting us access to their TPUs.

\bibliography{anthology,custom}
\bibliographystyle{acl_natbib}
\clearpage
\appendix

\section{Training Details}
\label{app:pretraining}
\subsection{Pre-training Details}
We pre-train the monolingual models for 50 epochs on the Internet Archive corpus and continue pre-training for 100 epochs on the born-digital texts, the trilingual models were trained for 100 epochs on the born-digital texts. The tokenizers were trained on the born-digital data only. \greBERTa and \philberta were trained on an NVIDIA A100-PCIE-40GB, \greTa and \philta on a Google TPU v2-8. Training took between 3 and 7 days. Further details in \Cref{tab:pretrainingdetails}.

\begin{table}
\centering
\resizebox{\linewidth}{!}{
\begin{tabular}{lllll}
\theadstart
    \textbf{Hyperparameter} &
  \textbf{\greBERTa} &
  \textbf{\philberta} &
  \textbf{\greTa} & 
  \textbf{\philta}\\
\tbody
Adam $\epsilon$ & $1 \cdot 10^{-8}$ &$1 \cdot 10^{-8}$ & $1 \cdot 10^{-8}$& $1 \cdot 10^{-8}$\\
Adam $\beta_1$ & $0.9$ & $0.9$&$0.9$&$0.9$\\
Adam $\beta_2$ & $0.999$ &$0.999$ &$0.999$&$0.999$\\

Attention Dropout & $0.1$ &$0.1$  & $0.1$ & $0.1$\\
Attention Heads &$12$ & $12$  & $12$& $12$\\
Batch Size & $128$ & $256$ & $512$ & $512$\\
$d_{\text{ff}}$ & --- & --- & $2048$ & $2048$\\
$d_{\text{kv}}$ & --- & --- & $64$ & $64$\\
$d_{\text{model}}$ & --- & --- & $768$ & $768$\\
Hidden Dropout & $0.1$ & $0.1$ & $0.1$ & $0.1$\\
Hidden Size &$768$ & $768$ &  ---& ---\\
Learning Rate (LR) & $5 \cdot 10^{-5}$ &  $5 \cdot 10^{-5}$&  $5 \cdot 10^{-3}$& $5 \cdot 10^{-3}$\\
LR Scheduler & linear & linear & linear & linear\\
Nb.\ of Layers & $12$& $12$ &  $2 \cdot 12$ & $2 \cdot 12$\\
Nb.\ of Parameters & $126$ mill. & $135$ mill.&$223$ mill.& $247$ mill.\\
Train Epochs & $50$, $100$& $0$, $100$ &  $50$, $100$& $0$, $100$\\
Warmup Steps & $0$ & $0$ & $10000$ & $10000$\\
Weight Decay & $0$& $0$ & $0.01$ & $0.01$\\
 \tend
\end{tabular}
}
\caption{Pre-training hyperparameters.}
\label{tab:pretrainingdetails}
\end{table}

\subsection{Fine-tuning Details}
We train every Greek model for 50 epochs on an NVIDIA GeForce GTX 1080 Ti, evaluating the model after every epoch on the validation set and using early stopping with a stopping patience of 5. As the EvaLatin dataset does not provide a validation set, we use 2\% of the training data as the validation set. Furthermore, since the EvaLatin dataset is larger than the Greek datasets, we set the maximum number of training epochs to 20 for the Latin models. Depending on the treebank and the task, training the models took approximately 1 hour (PoS tagging), 5--7 hours (dependency parsing), and 1--3 days (lemmatization). Further details in \Cref{tab:finetuningdetails}. We did not experiment with different hyperparameter settings, as our main goal was to provide comparable and wide-ranging benchmarking results. 

\begin{table}[t]
\centering
\resizebox{0.6\linewidth}{!}{
\begin{tabular}{ll}
\theadstart
    \textbf{Hyperparameter} &
    \\
\tbody
Adam $\epsilon$ & $1 \cdot 10^{-8}$\\
Adam $\beta_1$ & $0.9$ \\
Adam $\beta_2$ & $0.999$ \\
Batch Size & $32 $\\
Early Stopping Patience & $5$\\
Learning Rate & $1 \cdot 10^{-4} $\\
Learning Rate Scheduler & linear \\
Random Seeds & $42,1,2$\\
Train Epochs & $50$\\
Weight Decay & $1 \cdot 10^{-5}$ \\
 \tend
\end{tabular}
}
\caption{Fine-tuning hyperparameters.}
\label{tab:finetuningdetails}

\end{table}

\section{Downstream Task Details}
\label{app:datadetails}

\subsection{Universal Dependencies and EvaLatin 2022}
\label{app:udandevalatin}
\begin{table}[ht!]
\centering
\resizebox{0.9\linewidth}{!}{
\begin{tabular}{llll}
\theadstart
    &\textbf{Perseus}& \textbf{PROIEL} & \textbf{EvaLatin}\\
\tbody
Sentences (train) & \num{11476} & \num{15014} & \num{15785}\\
Sentences (dev) & \num{1137} & \num{1019} & ---\\
Sentences (test) & \num{1306} & \num{1047} & \num{1960
}\\
Sentences (total) & \num{13919} & \num{17080} & \num{17745}\\
Tokens (train) & \num{159895} & \num{187033}& \num{316573}\\
Tokens (dev) & \num{22135} & \num{13652}& ---\\
Tokens (test) & \num{20959} & \num{13314}& \num{45544}\\
Tokens (total) & \num{202989} & \num{213999}& \num{362117}\\
Lemmata & \num{13413} & \num{9348}& \num{10357}\\
Forms & \num{41304} & \num{32591}& \num{54133}\\
UPoS Tags & \num{14} & \num{14}& \num{16} \\
XPoS Tags & \num{847} & \num{27}& ---\\
Dependency Relations & \num{25} & \num{33} & ---\\
 \tend
\end{tabular}
}
\caption{Statistics of the Perseus, PROIEL, and EvaLatin datasets.}
\label{tab:datasetstatistics}
\end{table}

For PoS tagging, UD provides universal PoS tags (UPoS) and language-specific PoS tags (XPoS). UPoS consists of 17 tags used for all languages covered by UD.\footnote{In the case of AG, 3 of these 17 tags are not used.} XPoS tags, on the other hand, can follow a dataset-specific annotation scheme.  While the XPoS tags of the PROIEL dataset are similar to the UPoS tags, the Perseus dataset aims for a complete morphological analysis (cf.\ \Cref{subsec:methodology}). 

See \Cref{tab:datasetstatistics} for further details and \Cref{tab:overview} for an overview. In line with common convention, we report the accuracy for both PoS tag sets.
For dependency parsing, we report the unlabeled attachment score (UAS) and the labeled attachment score (LAS). The UAS indicates the percentage of tokens that have been assigned the correct head, whereas for the LAS, both the predicted head and the dependency label have to be correct.
All results are obtained from the official evaluation script.\footnote{\url{https://universaldependencies.org/conll18/conll18_ud_eval.py} and \url{https://github.com/CIRCSE/LT4HALA/blob/master/2022/data_and_doc/conll18_ud_eval_EvaLatin_2022_rev2.py}.}

\subsection{Semantic and World Knowledge}
\label{app:datasetcreation}

\paragraph{Semantic Knowledge.} We asked a graduate student and a doctoral candidate in the field of Classics to gather synonym and antonym pairs. Such word pairs can be nouns and substantivized adjectives or substantivized infinitives. We then utilized a predefined template to generate sentences that incorporate the collected pairs. As this template does not ensure grammaticality, the annotators manually edited the sentences. Subsequently, the sentences were independently reviewed by both annotators, deduplicated, and then verified by a professor of Ancient Greek. All three annotators participated on a voluntary basis and were not compensated for their contributions. One of the annotators is also a co-author of this paper.

141 synonym and 146 antonym pairs were collected. While we publish all 287 examples, we drop 5 randomly selected antonym pairs in our experiments to ensure that the number of synonym and antonym pairs is equal. We train all language models for 10 epochs using a batch size of 4 and report the averaged, cross-validated results.

\paragraph{World Knowledge.} This dataset was compiled by one of the previous annotators who is not a co-author of this paper. The annotator gathered 228 examples with 11 different relations by reading through Hesiod's \textit{Theogony} and by drawing inspiration from \citet{kern2003lexikon}, a lexicon that contains comprehensive information about mythical figures.

\section{Acquisition of Pre-training Data}
\label{app:corpuscreation}

\subsection{Ancient Greek Pre-training Data}
\paragraph{Open Greek \& Latin Project.\footnote{\url{https://opengreekandlatin.org/}.}} The Open Greek \& Latin Project is an umbrella project covering various subprojects that aim toward the development of open-access corpus linguistic resources for Latin and Classical Greek. Two of them, the Perseus Digital Library and the First Thousand Years of Greek project, contain Ancient Greek texts, mostly covering texts that are typically associated with classical antiquity, such as Homer, Plato, Herodotus, Euripides, and Plutarch. Already in this corpus, we find a wide variety of dialects and language stages.
The Open Greek \& Latin Project contains approximately $30$ million tokens.

\paragraph{Greek Medieval Texts.\footnote{\url{https://inventory.clarin.gr/corpus/890}.}} The   Greek Medieval Texts corpus offered by \textsc{CLARIN} covers writings from the fourth to the sixteenth century AD.  It contains religious, poetical-literary and political-historical texts as well as hymns and epigrams. Strictly speaking (and as the name suggests) the corpus contains texts of late antiquity,  
and in particular, Medieval Greek. 
We argue, however, that Ancient Greek and Medieval Greek, although different language stages, are strongly connected to each other and that our language models benefit from seeing more diverse data during pre-training. This corpus contains about $3.3$ million tokens and is licensed under the CC BY-NC 4.0 license.

\paragraph{Patrologia Graeca.\footnote{\url{http://patristica.net/graeca/}.}}

The Patrologia Graeca is a large collection of important Christian texts written in Greek, 
dating from the first until the fifteenth century AD. Since not all texts are machine-readable and  available,
we are restricted to those out of copyright texts that are made accessible 
(around $28.5$ million tokens).

\paragraph{Internet Archive.\footnote{\url{https://archive.org/}.}} The Internet Archive is an online library that
provides texts obtained from public domain books via
OCR.
We found out that only a small fraction of the books containing Ancient Greek text was labeled as such. Moreover, we discovered that even less books were transcribed with OCR settings that allowed Greek characters. As a result, many high-quality scans of Ancient Greek texts were transcribed into incomprehensible sequences of non-Greek characters. For example, the verse \textgreek{ὦ γύναι ἦ μάλα τοῦτο ἔπος νημερτὲς ἔειπες}\footnote{Hom. \textit{Il}. 3.204.} is transcribed as \texttt{\&  yvvai, ff pdXa tovto Stto\^{} vrjpepTe\^{} e\texteuro/.7r\texteuro9$\ast$}.

Even though transcriptions of this nature may seem useless at first glance, they are nevertheless helpful in identifying documents that have been incorrectly treated as non-Greek texts, for many common words are relatively consistently transcribed. \textgreek{τοῦτο} (\enquote{this}), for example, is often transcribed into \texttt{tovto}. By searching for all books that contain the word \texttt{tovto}, we can identify potential Greek texts. This approach allows us to avoid the computationally intensive task of applying Greek OCR to every book in the Internet Archive, and instead focus our efforts on a more targeted search. All candidates are then filtered more aggressively: If a candidate contains the five (presumably) Greek stopwords \texttt{tovto} (\textgreek{τοῦτο}), \texttt{kal} (\textgreek{καί}), \texttt{tov} (\textgreek{τόν}), \texttt{to} (\textgreek{τό}), and \texttt{yap} (\textgreek{γάρ}) more than ten times, the candidate is considered to contain Greek text.

We argue that this method effectively minimizes false positives while retaining a high recall: Since Greek stopwords like \textgreek{τοῦτο} (\enquote{this}) and \textgreek{καί} (\enquote{and}) should be present often enough in every book with a significant amount of text, our approach should correctly classify them as Greek. Non-Greek texts, on the other hand, should hardly contain all five stopwords more than ten times.

This procedure yields $25378$ books, on which the Internet Archive applies OCR with Ancient Greek as a target language. While our method reliably detects Greek texts, it does not ensure a high scan (and therefore also text) quality. In order to use solely high-quality data, we keep only lines in which more than $90$\% of tokens are also present in the born-digital vocabulary. A similar approach is used by \citet{bamman2020latin}, who use Latin texts from the Internet Archive as pre-training data for Latin \bert. They \enquote{retain only those books where at least $40$\% of tokens are present in a vocabulary derived from born-digital texts}.
We argue that it is more sensible to include or disregard individual lines instead of whole books: Almost every Greek text contains a Latin or English introduction, and many books are equipped with a translation. Thus, our method not only ensures high-quality data but also removes non-Greek text parts.

Finally, to ensure that our dataset does not contain any significant duplications, we remove all instances of repeated text exceeding 300 characters. After this aggressive filtering, we have approximately 123.3 million tokens left. To demonstrate its quality, we show 40 samples randomly drawn from the dataset in \Cref{tab:samples}.

\begin{table}[ht!]
\centering
\small{
\begin{tabular}{@{}p{7cm}@{}}
\tbody
\textgreek{τῆς Μασίστεω γυναικός, ἐούσης καὶ ταύτης ἐνθαῦτα. ὡς
}\\
\textgreek{πίστις ὑμῶν; φοβηθέντες δὲ ἐθαύμαζον, λέγοντες πρὸς ἀλλήλους,
}\\
\textgreek{ὑποληπτέον" ἡ γὰρ πέψις τοῖς μὲν ἐν τῷ ἄνθει μᾶλλον
}\\
\textgreek{ἀνέπαυσαν γὰρ τ. ἐμὸν πνεῦμα κ. τὸ ὑμῶν
}\\
\textgreek{εἰ Σατανᾶς ἀνέστη ἐφ᾽ ἑαυτόν
}\\
\textgreek{πρόσωπον ναοῦ Κυρίου, μετὰ τὸ ἀποικίσαι Ναβουχοδονόσορ
}\\
\textgreek{ἐκείνοις δὲ ὄντος ἀεὶ τοῦ ἐπιχειρεῖν καὶ ἐφ ἡμῖν εἶναι δεῖ τὸ προαμύνασθαι.
}\\
\textgreek{ἑξακοσίους ὁπλίτας εἰς τὴν πόλιν ἄ ἄγει. ἐν τῷ στρατεύματι
}\\
\textgreek{ἔχοντι τοῦ Γερμανικοῦ συναγορεύειν μέλλοντος,"
}\\
\textgreek{νοοῦν εἴη, ὅτι ἄλλου τὴν νόησιν λαμβάνον οὐ τὸ
}\\
\textgreek{ἐὰν δὲ μὴ τούτοις δύνῃ χρῆσθαι,
}\\
\textgreek{μου- ἐφ᾽ ὑμᾶς" ὑμεῖς.δὲ καθίσατε ἐν τῇ πόλει Υ Ἱερουσαλὴμ
}\\
\textgreek{καὶ νοητῆς τελειότητος.
}\\
\textgreek{μένον οὐκ ἐπίστευσαν.
}\\
\textgreek{τίον ἀράτω Ἰησοῦς
}\\
\textgreek{διδόντα ὑετὸν ἐπὶ τὴν γῆν, ἀποστέλλοντα ὕδωρ
}\\
\textgreek{ταρασσέσθω ὑμῶν ἡ καρδία, μηδὲ δευλιάτω. ἠκούσατε ὅτι ἐγὼ
}\\
\textgreek{τὴν ξημίην ἐπέθηκαν. Ζυώδεκα δέ μοι δοκέουσι πόλιας ποιή-
}\\
\textgreek{ἐστι’ τὰ δὲ γενητὰ, ἔξωθεν ὄντα, πρόσκειται, ὡς τῇ
}\\
\textgreek{ὁ δὲ Κλεομένης τὸν ἱρέα ἐκέλευε τοὺς εἵλωτας ἀπὸ τοῦ
}\\
\textgreek{ἅπαξ ἀλλὰ πολλάκις.
}\\
\textgreek{ἐλθόντος. καὶ αὖθις ἔδοξε τούτου χάριν καὶ
}\\
\textgreek{κερματίξῃς αὐτό, ἐκεῖνοι πολλαπλασιοῦσιν, εὐλαβούμενοι
}\\
\textgreek{καὶ προλάμψαν τὸ ἐραστὸν αὐτοῦ καὶ τὸν κρυπτόμενον
}\\
\textgreek{πεντακοσίας, οὺς πάντας ἡ τοῦ δεσπότου χάρις καὶ φιλανθρωπία διατρέφει.
}\\
\textgreek{ταύτης ἰδίᾳ προετρέπετο τὸν Σικόπαν κοινωνῆσαι
}\\
\textgreek{οὐδὲ παναρμονίου ἡμῖν δεήσει ἐν ταῖς ὠδαῖς τε καὶ
}\\
\textgreek{σημεῖα τοῦ τοῦτον συχοφαντεῖν ἐγχαλοῦντ᾽ ἀφορμήν.
}\\
\textgreek{συμπεριλαμβανομένων καὶ ψυχῆς καὶ τῶν ἐν
}\\
\textgreek{πλὴν ἐξ ὠκυβόλων εἴ ποτε τόξων
}\\
\textgreek{σφι ἄρτισις περὶ τὸ σῶμα ἐστί.
}\\
\textgreek{μὴ πέσῃς κατέναντι ἐνεδρεύοντος
}\\
\textgreek{ο Εἰς τοῦτο ἐσυνέργησαν οἱ πρῶτοι τῆς γενεᾶς τῆς,
}\\
\textgreek{χωρίων ἢ οἰκιῶν ὑπῆρχον, πωλοῦντες ἔφερον τὰς τιμὰς
}\\
\textgreek{ᾧ δὲ περὶ ἑκάστην μεθοδον") φιλοσοφοῦντι καὶ μὴ"
}\\
\textgreek{τῶν τῆς. παιδὸς γάμων, Ζεὺς διαλύσας ἐπέτρεψεν
}\\
\textgreek{ὑμῶν. πόλεις αἱ πρὸς νότον συνεκλείσθησαν, καὶ οὐκ ἦν ὁ ἀνοίγων: ἀπῳκίσθη Ιουδας,
}\\
\textgreek{πειρασμούς. Περὶ ταύτης ἡ Γραφὴ (ά. Κορ,
}\\
\textgreek{ἔπεσεν ἐπὶ πρόσωπον αὐτοῦ προσευχόμενος
}\\
\textgreek{ζητεῖ" οἷδεν. γὰρ ὁ-πατὴριὑμῶν ὁ οὐράνιος
}\\
 \tend
\end{tabular}
}
\caption{40 randomly drawn lines of the Internet Archive pre-training dataset.}
\label{tab:samples}
\end{table}

\subsection{English Pre-training Data}
By collecting English translations of ancient texts, we focus on texts that are strongly connected to antiquity. We gather these texts from various sources: The Perseus Digital Library\footnote{\url{http://www.perseus.tufts.edu/hopper/}.} and the Internet Classics Archive\footnote{\url{http://classics.mit.edu/}.} provide born-digital open-access translations of Classical Greek and Latin texts. Similarly, the Documenta Catholica Omnia database\footnote{\url{http://www.documentacatholicaomnia.eu/}.} contains a large amount of primarily catholic texts in many languages, of which we select the English partition for our use. Finally, we utilize Lexundria,\footnote{\url{https://lexundria.com/}.} Loebulus,\footnote{\url{https://ryanfb.xyz/loebolus/}.} and the Project Gutenberg to add (often proofread) scans of books in the public domain. While Lexundria and Loebulus are restricted to providing translations of Latin and Greek texts, the Project Gutenberg offers a more diverse range of literature. Therefore, we use only books from Project Gutenberg that are tagged with the keyword \enquote{Latin}.
We report detailed statistics in \Cref{tab:statisticscomplete}.

\begin{table}[ht!]
\centering
\resizebox{\linewidth}{!}{
\begin{tabular}{llr}
\theadstart

\textbf{Language} &
\textbf{Dataset} &
   \textbf{Number of Tokens}\\
\tbody

  \multirow{5}{*}{Ancient Greek} & Open Greek \& Latin & \hphantom{0}$30.0$ million\\
  &Greek Medieval Texts & \hphantom{00}$3.3$ million\\
   &Patrologia Graeca & \hphantom{0}$28.5$ million\\
   &Internet Archive & $123.3$ million\\
   &Overall& \textbf{185.1 million}\\
   \hdashline
   Latin & Corpus Corporum & \textbf{167.5 million}\\
   \hdashline
   \multirow{7}{*}{English} & Perseus &$10.8$ million\\
   &Classics Archive & $4.9$ million\\
   & Lexundria & $2.8$ million\\
   & Loebulus & $14.0$ million\\
   & Project Gutenberg & $28.7$ million\\
   & Documenta Catholica Omnia & $151.7$ million\\
   & Overall & \textbf{212.8 million}\\
   
 \tend
\end{tabular}}
\caption{Statistics of the pre-training datasets. Only Open Greek \& Latin is used by \citet{singh-etal-2021-pilot} and \citet{plutarchsshadows} for their
AG \bert models. Token counts 
determined by UNIX command \texttt{wc -w}.}

\label{tab:statisticscomplete}
\end{table}

\section{Further Results}
\label{app:furtherresults}

\begin{table}[ht!]
\centering
\begin{tabular}{ll}
\theadstart
      \textbf{Model} &
     \textbf{Accuracy}\\
\tbody
\cltk & $96.51$\\
\cdashline{1-2}
\udpipe (official) & $94.71$\\
\udpipe (ours) & $93.87$ $(0.05)$\\
\udpipe + \greBERTa & $94.17$ $(0.05)$ \\
\greTa & $97.40$ $(0.02)$\\
\greTa + Chars & $\mathbf{97.48}$ $\mathbf{(0.02)}$\\
\tend
\end{tabular}
\caption{Lemmatization results on the Ancient Greek PROIEL dataset. The results are averaged
over three runs with different random seeds, and the standard deviation is indicated in parentheses, except for the \cltk and \udpipe (reported results).}
\label{tab:lemmatizationproiel}
\end{table}

\begin{table*}[ht!]
\centering
\resizebox{0.8\linewidth}{!}{
\begin{tabular}{lllll}
\theadstart
   \cellcolor{gray!25} \textbf{Model} &
    \multicolumn{2}{c}{\cellcolor{gray!25}\textbf{PoS Tagging}} &
    \multicolumn{2}{c}{\cellcolor{gray!25}\textbf{Dependency Parsing}}\\
 \tsubheadstart \cellcolor{gray!25}
 &
    \cellcolor{gray!25}\tsubhead UPoS &
   \cellcolor{gray!25} \tsubhead XPoS &
    \cellcolor{gray!25}\tsubhead UAS & 
    \cellcolor{gray!25}\tsubhead LAS 
    \\
\tbody
\cltk & $97.10$ & $97.47$ & $76.81$ & $73.39$\\
\cdashline{1-5}
\udpipe (official) & $97.77$ & $98.05$& $86.05$& $82.14$\\
\udpipe (ours) & $97.99$ $(0.05)$ & $97.68$ $(0.06)$& $85.64$ $(0.17)$ & $81.70$ $(0.25)$\\ 
\udpipe + \greBERTa & $98.56$ $(0.02)$ & $\mathbf{98.70}$ $\mathbf{(0.03)}$& $89.75$ $(0.16)$& $86.59$ $(0.15)$\\ 
    AG \bert \cite{singh-etal-2021-pilot} & $97.98$ $(0.02)$ & $98.14$ $(0.05)$& $88.50$ $(0.09)$ &$84.72$ $(0.18)$\\
    AG \textsc{BERT} \cite{plutarchsshadows}& $97.19$ $(0.06)$& $97.42$ $(0.08)$& $86.61$ $(0.21)$& $82.12$ $(0.15)$ \\
    \gretaenc & $98.16$ $(0.02)$ & $98.31$ $(0.03)$& $89.93$ $(0.08)$& $86.48$ $(0.08)$\\
    \philberta & $98.15$ $(0.16)$ & $98.45$ $(0.05)$ & $\mathbf{90.32}$ $\mathbf{(0.13)}$& $86.43$  $(0.61)$\\
    \greBERTa & $\mathbf{98.60}$ $\mathbf{(0.03)}$& $\mathbf{98.70}$ $\mathbf{(0.04)}$  & $90.28$ $(0.03)$& $\mathbf{86.84}$ $\mathbf{(0.12)}$\\
 \tend
\end{tabular}
}
\caption{PoS tagging and dependency parsing results on the Ancient Greek PROIEL dataset. The results are averaged over three runs with different random seeds, and the standard deviation is indicated in parentheses, except for the \cltk and \udpipe (reported results). Note also that the \cltk is not trained on exactly the same data as the other models and therefore not directly comparable.}
\label{tab:posanddepsproiel}

\end{table*}

\begin{figure}[!th]
    \centering
    \includegraphics[width=\linewidth]{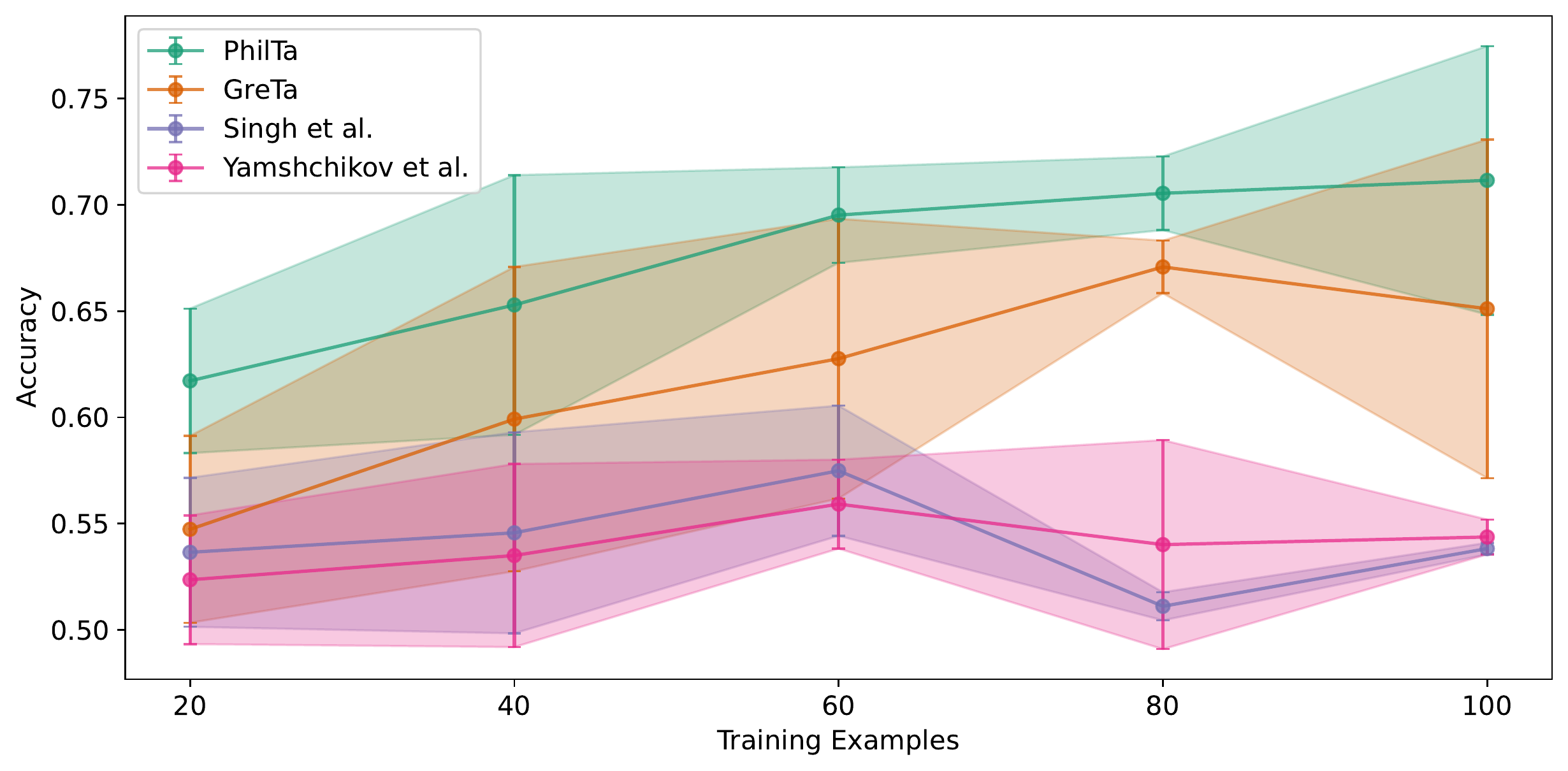}
    \caption{Synonym/antonym disambiguation accuracy scores for different few-shot training set sizes, for \greTa and \philta against \agbert models. The models are always given 
equal amounts of synonyms and antonyms, e.g., when using $20$ training instances, the models are given 
$10$ synonyms and $10$ antonyms. 
We evaluate all models using $k$-fold cross-validation and report standard deviation as error bars.}
    \label{fig:synonymsencdec}
\end{figure}

\end{document}